\documentclass{article}
\usepackage{acra}

\usepackage{booktabs}
\usepackage{placeins}
\usepackage{graphicx}
\usepackage{subfig}
\usepackage{hyperref}

\title{\LARGE \bf
A Service Robot's Guide to Interacting with ``Busy'' Customers 
}

\author{Suraj Nukala$^{1}$, Meera Sushma$^{1}$, Leimin Tian$^{2}$, Akansel Cosgun$^{3}$ and Dana Kuli\'{c}$^{1}$%
\thanks{$^{1}$Suraj Nukala, Meera Sushma, and Dana Kuli\'{c} are with Faculty of Engineering, Monash University, Melbourne, VIC 3800, Australia
        {\tt\small snuk0002@student.monash.edu}, 
        {\tt\small msus0012@student.monash.edu}, 
        {\tt\small dana.kulic@monash.edu}}
\thanks{$^{2}$Leimin Tian is with CSIRO Robotics
        {\tt\small Leimin.Tian@csiro.au}}
\thanks{$^{3}$Akansel Cosgun is with the School of Information Technology, Deakin University
        {\tt\small Akansel.Cosgun@deakin.edu.au}}%
}

\begin{document}

\maketitle
\pagestyle{empty}

\begin{abstract}

The growing use of service robots in hospitality highlights the need to understand how to effectively communicate with pre-occupied customers. This study investigates the efficacy of commonly used communication modalities by service robots, namely, acoustic/speech, visual display, and micromotion gestures in capturing attention and communicating intention with a user in a simulated restaurant scenario. We conducted a two-part user study (N=24) using a Temi robot to simulate delivery tasks, with participants engaged in a typing game (MonkeyType) to emulate a state of busyness. The participants' engagement in the typing game is measured by words per minute (WPM) and typing accuracy. In Part 1, we compared non-verbal acoustic cue versus baseline conditions to assess attention capture during a single-cup delivery task. In Part 2, we evaluated the effectiveness of speech, visual display, micromotion and their multimodal combination in conveying specific intentions (correct cup selection) during a two-cup delivery task. The results indicate that, while speech is highly effective in capturing attention, it is less successful in clearly communicating intention. Participants rated visual as the most effective modality for intention clarity, followed by speech, with micromotion being the lowest ranked.These findings provide insights into optimizing communication strategies for service robots, highlighting the distinct roles of attention capture and intention communication in enhancing user experience in dynamic hospitality settings.
\end{abstract}

\begin{figure}
    \centering
    \includegraphics[width=0.75\linewidth]{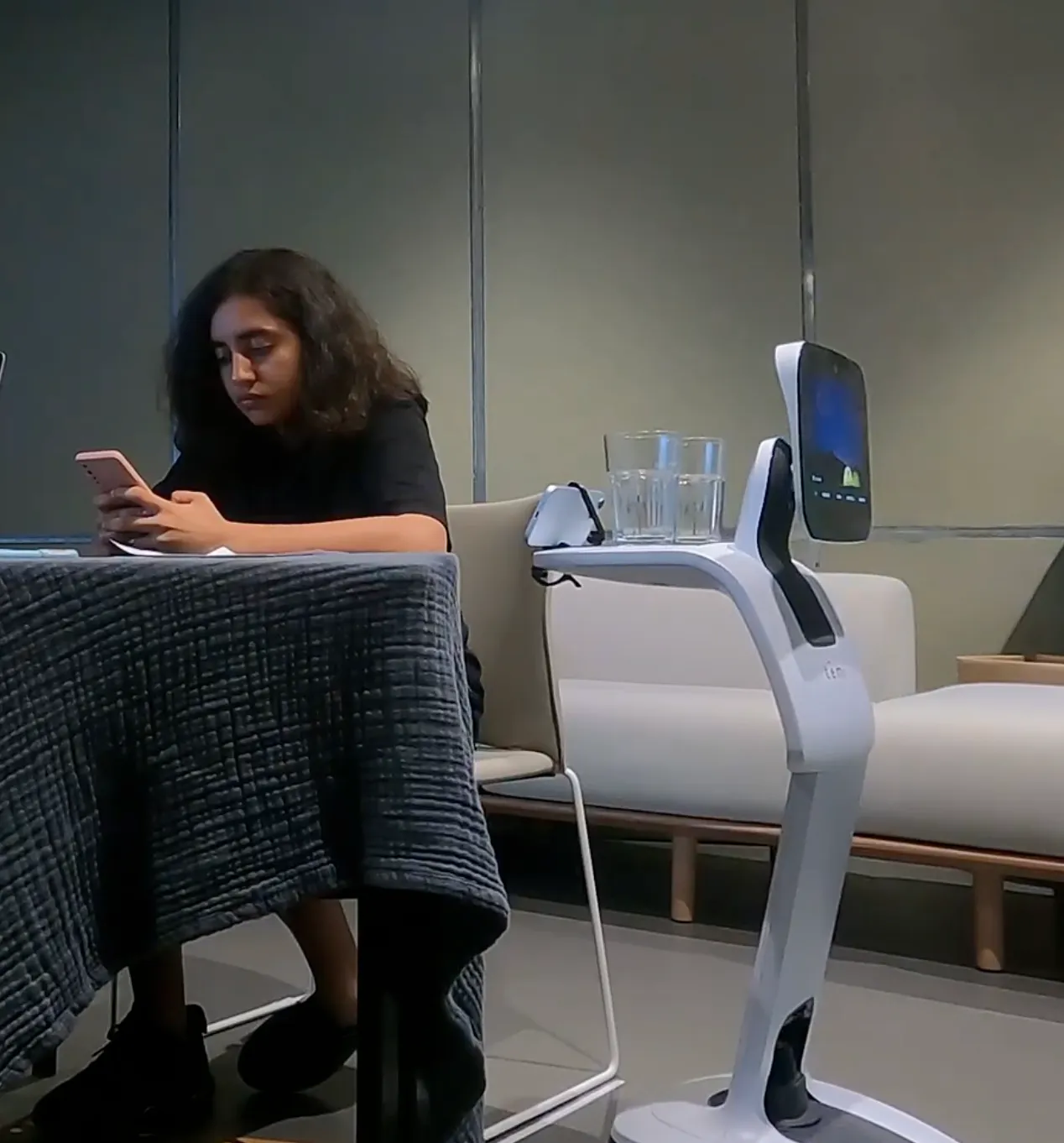}
    \caption{User study scenario (intention communication stage): the Temi service robot is delivering an order to a ``busy'' customer, who needs to pick up the correct glass from the two on its tray while playing a typing game.}
    \label{fig:enter-label}
\end{figure}

\section{INTRODUCTION}

Service robots are increasingly deployed in restaurant and cafe environments, where they need to effectively communicate intent to customers who are often preoccupied with cognitively demanding tasks~\cite{c1}, \cite{c13}. Currently, customers are excited to interact with and test the capabilities of these novel robotic systems. However, as service robots become commonplace in the near future, their novelty will fade, and customers are expected to prioritize personal tasks over robot interactions~\cite{c13}. A critical challenge in such ``in-the-wild'' deployments is the **lack of social licence**---established norms governing acceptable robot behavior remain absent, unlike human--human interactions where implicit rules (e.g., yielding in hallways) are culturally ingrained~\cite{c16}. Without these norms, users cannot reliably predict robot actions, increasing hesitation, miscommunication, and collision risk in shared spaces. To simulate such attention shift, our study involves participants playing a typing game, MonkeyType\footnote{\url{https://monkeytype.com/}}, on their phones, creating a distracted state that challenges robots to deliver clear order indications. Prior work has explored single-modality cues for intent communication, including visual displays~\cite{c2}, \cite{c4}, \cite{c6}, motion-based signals~\cite{c8},\cite{c12}, and verbal feedback~\cite{c9}, \cite{c10}. For example,~\cite{c4} showed that projected trajectories speed human responses, while~\cite{c10} found projections aid verbal disambiguation. Yet single modalities often fail to grab the attention of busy customers. To address this, we leverage sense of agency (SoA)—the user’s feeling of control over outcomes—and intentionality—seeing the robot as acting with purpose, not mechanically.~\cite{c11} show that perceiving robot intentionality helps users predict actions, boosting SoA even when distracted. Multimodal cues (micromotion, visualization, speech) make intent more salient, strengthening SoA and enabling quick, confident responses during primary tasks.Few studies have investigated multimodal approaches for order indication. This study investigates the effectiveness of a multimodal approach (combining micromotion, visualization, and speech) for robot interactions with users performing cognitively demanding tasks. We hypothesise that multimodal communication will achieve higher order delivery accuracy than any single modality alone by improving intentionality attribution and interaction clarity. We evaluate the baseline (no communication), micromotion, visualization, speech, and multimodal combination in a user study to evaluate their effectiveness in enhancing order accuracy and user experience.

\section{RELATED WORKS}

\subsection{Visual and Motion-Based Intent Communication}
\label{subsec:visual_motion_intent}
Visual and motion-based cues are critical for robots to communicate intent in human-robot interaction (HRI), particularly for tasks like order indication. ~\cite{c1} demonstrated that visual and movement cues from a robot waiter effectively convey intent to customers, enhancing interaction clarity. ~\cite{c2} found that augmented reality (AR) visualizations, combining robot intent and pedestrian path predictions, significantly improved trust in co-navigation scenarios. Similarly, ~\cite{c4} showed that projecting a robotic forklift's trajectory and occupied space improved human response times and safety perceptions. ~\cite{c6} confirmed AR's efficacy in object handovers, particularly under error conditions, outperforming non-AR conditions. ~\cite{c8} identified projected arrows as socially acceptable motion legibility cues compared to flashing lights in a user study. ~\cite{c16} further revealed that in narrow hallway navigation, novice users strongly prefer explicit text- and speech-based instructions (``Please continue moving forwards'') over intent signals (``I will move out of the way''), while experienced users favor the less intrusive intent-based approach, suggesting adaptive communication for real-world concierge robots. These findings highlight the effectiveness of visual and motion cues, akin to our study's use of visualization and micromotion for order indication. However, these existing works assumed that the robot will have the full attention of the user during the interaction, which may not accurately represent the service application scenario where the user may be preoccupied with other tasks and receiving orders from the robot is only a secondary task, especially as robots become commonly adopted. We extend existing research by integrating visual and motion-based cues with speech in a multimodal framework, investigating how individual modalities and their multimodal combination enhances order accuracy for users preoccupied with tasks.

\subsection{Object Handover Coordination}
\label{subsec:handover_coordination}
Effective object handovers in HRI rely on coordinated cues and user-centric configurations, relevant to our order indication task involving physical transfers.~\cite{c3} analyzed human-human handovers, identifying physical and verbal cues that robots can adopt to streamline object transfers.~\cite{c5} advocated incorporating user preferences in handover configurations, emphasizing object visibility, default orientation, and affordances for natural interactions.~\cite{c7} found that a minimum-jerk velocity profile at 0.225 m/s was perceived as safe and efficient for robot-human handovers.~\cite{c14} reviewed robotic handover strategies, underscoring the importance of context and user-centric design. These studies inform our use of micromotion, visualization, and speech as coordinated cues for order indication. Unlike prior work focusing on single modalities or specific handover mechanics, our multimodal approach aims to enhance accuracy by leveraging multiple cues, potentially improving user awareness during a handover task while users are engaged in a concurrent task.

\subsection{User Experience and Preferences in HRI}
\label{subsec:user_experience}
User experience metrics, such as trust and perceived intelligence, are pivotal in designing HRI systems that align with user preferences.~\cite{c12} tested motion-based and light-based intent signals, finding that an LED bracelet near the robot's end effector was the most noticeable and least confusing for users sorting colored blocks, emphasizing user comfort.~\cite{c13} reported that users prefer robots approaching from the side rather than the front, as direct approaches can feel threatening, informing non-intrusive cue design.~\cite{c15} developed instruments to measure anthropomorphism, likeability, and perceived intelligence, aligning with our evaluation of trust and satisfaction in response to multimodal cues. These studies underscore the importance of user-centric design in HRI, relevant to our hypothesis that a multimodal combination enhances user experience. Our work builds on this by combining micromotion, visualization, and speech, hypothesizing that this approach better meets user expectations for clear communication, especially when users are preoccupied with tasks, potentially improving perceived interaction quality.

\subsection{Verbal and Multimodal Intent Disambiguation}
\label{subsec:verbal_multimodal}
Verbal and multimodal feedback plays a key role in disambiguating robot intent, particularly in complex interaction scenarios.~\cite{c9} found that mixed reality (MR) feedback was 10\% more accurate than physical pointing in a bidirectional communication study, highlighting the potential of multimodal interfaces.~\cite{c10} showed that projections outperformed head-mounted displays for disambiguating verbal requests in a Lego block pickup task, with users preferring projections for clarity.~\cite{c11} introduced the vicarious sense of agency, suggesting that attributing intentionality to robots enhances action-outcome perceptions, critical for effective HRI. These studies align with our use of speech and multimodal cues (speech, micromotion, visualization) to clarify order indication. Unlike prior work focusing on single or dual modalities, our study combines three modalities, hypothesizing that this integration enhances order accuracy by making intent more salient to users, even when preoccupied with tasks, potentially strengthening intentionality attribution and interaction clarity.

\section{METHODOLOGY}

Given previous research on service robots for food delivery, this paper seeks to extend the existing knowledge by investigating the importance of the mode of communication in the effectiveness of a service robot. We measure objective service outcomes as the accuracy of the item being chosen correctly, the keyboard typing game metrics to measure focus, as well as subjective user experience and service satisfaction in a user study. The number of participants ($N=24$) was determined using G*Power Analysis. For both studies, we assumed a medium effect size ($f = 0.25$), $\alpha = 0.05$, and power $= 0.80$. The study protocols were reviewed and approved by the Monash University Human Research Ethics Committee (Project ID: 45951).

\subsection{Interaction scenarios}

\begin{figure}
    \centering
    \includegraphics[width=0.45\linewidth]{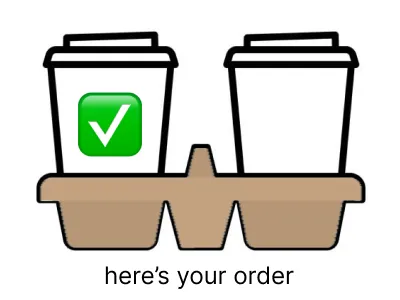}   
    \caption{Example visualisation in service intention communication: the user should pick up the cup on the left}
    \label{fig:vis} 
\end{figure}

\begin{figure}[h]
    \centering
      \includegraphics[width=\linewidth]{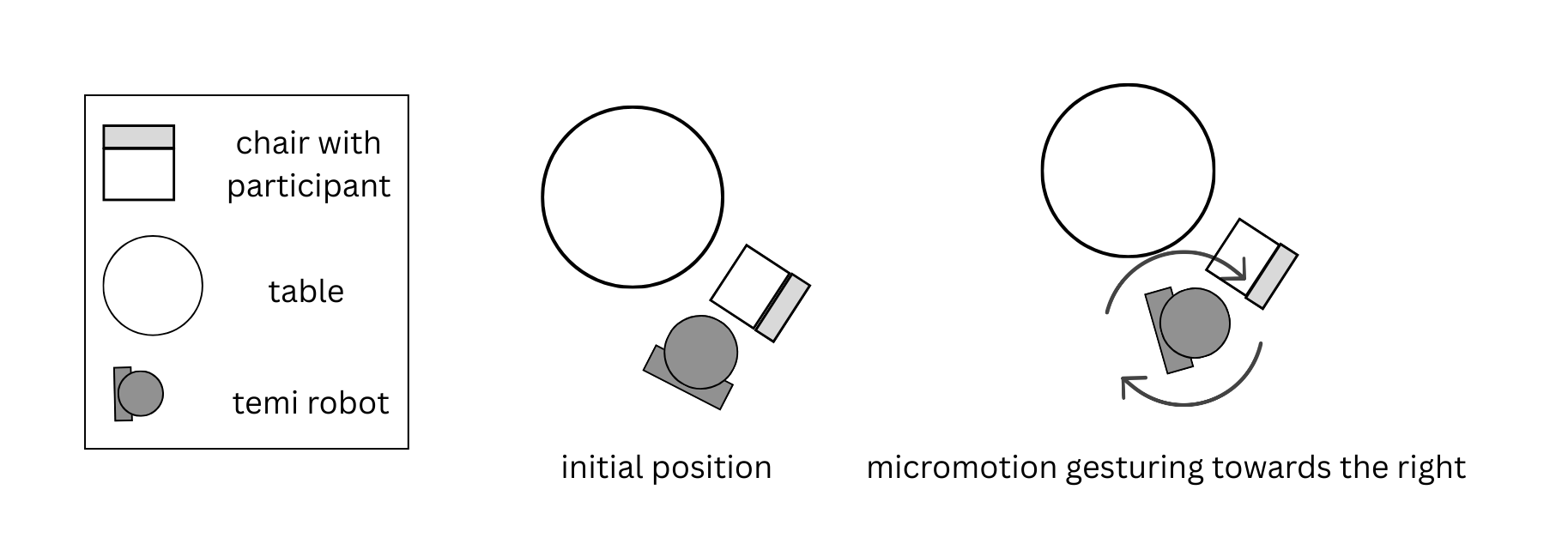}
      \caption{Illustration of micromotion for service intention communication: after the greeting gesture of slight bowing, the robot rotates to position the cup that the user should pick up to be closer}
      \label{fig:micromotion}
\end{figure}

We designed a 2-part user study to evaluate the effectiveness of various modalities for robot order indication with busy customers. We simulated service robot interaction with busy customers by asking participants to engage in a cognitively demanding task, playing the typing game MonkeyType on their phones. Prior to starting each part of the study and each round, the participants were briefed about the scenario and were instructed to select a cup based on perceived robot intent. Note that they were unable to reject the robot's service request and the session only progresses after a cup has been taken. To isolate the effect of communication modality on intent perception and reaction, participants were informed that the robot would arrive once per typing round at a consistent interval (45 seconds), ensuring expectation while varying only the cue type (acoustic, visual, motion, speech, or multimodal). They also completed a practice round to familiarize themselves with the game.  In Part 1 of the study, we compared a baseline condition without communication to a non-verbal acoustic robot communication to assess the impact of auditory cues on attracting user attention on robot's intention to deliver an order. The participants experienced the two deliveries in counter-balanced order. Then in Part 2 of the study, we compared visualisation, micromotion, speech, and a multimodal combination (all) of robot service intention communication to rank their effectiveness in enhancing order delivery accuracy and user experience with the same participant playing the same typing game. Note that before communicating the service intention, the robot used the same non-verbal acoustic cue in Part 1 to gain the user's attention. In each condition, the robot (Temi) performed specific actions to indicate the delivery of a cup. Each participant experienced the four deliveries with different robot communication modalities in random order to account for habituation. The robot’s behaviors are detailed below:

\begin{itemize}
 \item \textbf{Part 1: Baseline vs Non-verbal acoustic cues for user attention capturing}
    \begin{itemize}
        \item \textit{Baseline}: The robot approaches the participant with one cup, remains stationary for 10 seconds, and then departs without any additional cues. Participants may notice the robot in their peripheral vision, but, being preoccupied with the typing game, they may or may not pick up the cup, depending on their level of attention.
        \item \textit{Non-verbal acoustic cues}: The robot approaches with one cup and emits a distinctive sound upon arrival, sourced from the character EVA in the movie \textit{WALL-E}. This sound was chosen for the design similarity between EVA and Temi, enhancing the robot’s aesthetic coherence. After emitting the sound, the robot waits for 10 seconds and departs. We hypothesize that the sound alerts participants, increasing the likelihood of noticing the robot and interacting with the cup during or after their MonkeyType round.
    \end{itemize}
    Part 1 aims to determine the effectiveness of an non-verbal auditory cue in capturing the attention of distracted users compared to a baseline with no active indication.

\item \textbf{Part 2: Visualisation, Micromotion, Speech, and Multimodal intention communication}
    The robot approaches with two cups and emits the EVA sound same as Part 1 upon arrival and then based on the delivery round interacts with the participant with a different modality.
    \begin{itemize}

    \item \textit{Visualisation}: Upon arrival, the robot displays a visualization on its onboard screen indicating which of the two cups is the correct one for the participant to pick up with the text message ``here's your order'' (see Figure~\ref{fig:vis}). After 10 seconds, the robot departs. The visualization aims to provide a clear visual cue to guide user action, despite their engagement in MonkeyType.

    \item \textit{Micromotion}: Upon arrival, the robot performs a micromotion gesture by bowing slightly, a culturally respectful gesture in many contexts, to draw attention. It then executes a second micromotion by rotating itself to position the correct cup closer to the participant (see Figure~\ref{fig:micromotion}). After 10 seconds, the robot departs. These gestures aim to subtly signal the correct cup while respecting the participant’s focus.

    \item \textit{Speech}: Upon arrival, the robot uses its onboard text-to-speech engine to verbally instruct the participant to pick up the correct cup. After 10 seconds, it departs. The speech cue aims to provide explicit guidance.

    \item \textit{All (Multimodal)}: Upon arrival, the robot simultaneously performs all three actions: displaying the visualization, performing the micromotion gestures (bowing and rotating), and delivering the verbal instruction via text-to-speech. It waits for 10 seconds before departing. This condition tests the combined effect of all modalities to maximize saliency and clarity.
    \end{itemize}
    
Part 2 aims to investigate the effectiveness of each modality (visualisation, micromotion, speech) individually and evaluate the benefit of combining them in a multimodal approach for order accuracy and user experience for busy users.
\end{itemize}




\subsection{Evaluation metrics}

To investigate the influence of intention communication modalities on interaction outcomes and user experiences, we collected both objective and subjective measures. We recorded the order pick-up time in each delivery round in Part 1 and Part 2, as well as the delivery accuracy in each round in Part 2 (whether the correct cup was chosen or not) as objective measures of the interaction outcome. Additionally, we measured participants’ game scores (typing speed measured as word per minute and typing accuracy) to assess their level of focus during the interaction with Temi.

An evaluative questionnaire was used to collect participants’ subjective perceptions after each interaction. We used the  question from ~\cite{c14} for evaluating perception towards the robot, the human-robot collaboration questionnaire for evaluating the interaction fluency, human-robot trust, and working alliance. In addition, we included a few ad-hoc questions on service satisfaction and perceived service efficiency. All items were measured on a 5-point Likert scale. We have also performed a manipulation test, where after each interaction the participants answered if they think they picked up the correct cup and shed light on the modalities they perceived. The participants’ responses to the questionnaire, and interview summaries of the experiment have been provided in the accompanying dataset.
Further, we conducted exit interviews with each participant to better understand their interaction with the robot, their perception and understanding of Temi’s intentions and feedback as well as comparison to any prior experiences.

Here are the definitions of the metrics that we obtained:

\begin{itemize}
    \item \textbf{Game Metrics}
    \begin{itemize}
        \item {\textbf{Words Per Minute (WPM)}}: The WPM score from MonkeyType reflects the number of words typed.
        \item {\textbf{Game Accuracy Score}}: The accuracy score from MonkeyType indicates the percentage of typing errors or mistakes.
    \end{itemize}
    \item \textbf{Video Footage Metrics}
    \begin{itemize}
        \item {\textbf{Delivery Time}}: This measures the number of seconds the robot took to deliver the cup, from the start of the round until the participant picks the cup up.
        \item {\textbf{Focus (on Robot) Time}}: This tracks the number of seconds the participant focused on the robot (determined by gaze).
        \item {\textbf{Focus (on Game) Percentage}}: A ratio calculated as [delivery time - focus (on robot) time] divided by interaction time.
        \item {\textbf{Number of Distractions from Game}}: From video footage, this counts the instances where a participants' gaze shifted from the game to the robot.
    \end{itemize}
    \item \textbf{Pickup Accuracy Metrics}
    \begin{itemize}
        \item {\textbf{Cup Selection}}: From the survey, collects data from participants regarding whether they retrieved the cup.
        \item {\textbf{Percieved Accuracy}}: From the survey, assesses the participants' confidence in correctly picking the cup.
        \item {\textbf{Actual Accuracy}}: From the survey, evaluates the true accuracy of the participants' cup pick.
    \end{itemize}
    \item \textbf{User Experience Metrics}
    \begin{itemize}
        \item {\textbf{Work Fluency}}: From survey, measures how smoothly Temi and the participant worked together.
        \item {\textbf{Intelligence}}: From survey, gauges the perceived intelligence of Temi.
        \item {\textbf{Interaction Fluency}}: From survey, evaluates Temi's role in enhancing interaction fluency.
        \item {\textbf{Reliability}}: From survey, assesses the trustworthiness of Temi.
        \item {\textbf{Dependability}}: From survey, reflects confidence in Temi performing the right action at the right time.
        \item {\textbf{Goal Perception}}: From survey, examines how well Temi understood the participants' goals.
        \item {\textbf{Goal Communication}}: From survey, checks the participants' comprehension of Temi's objectives.
        \item {\textbf{Collaboration}}: From survey, evaluates if Temi and the participant shared aligned goals.
        \item {\textbf{Satisfaction}}: From survey, rates the participants' satisfaction with Temi's service.
        \item {\textbf{Speed}}: From survey, assesses how quickly Temi delivered the order.
    \end{itemize}
    
\end{itemize}

\subsection{Hypotheses}

Our main research question is ``how should robots interact with cognitively occupied customers?'' Specifically, we investigate the influence of the robot’s communication modality on the user perception for accuracy and clarity. Our hypotheses are:

\subsubsection{Part 1: Baseline vs non-verbal acoustic cue for attracting user attention}
\begin{itemize}

\item H1: The presence of a non-verbal acoustic cue while the robot approaches the user will improve user experience compared to having no cue. 

\end{itemize}

\subsubsection{Part 2: Visualisation, Micromotion, Speech, and
Multimodal intent communication}
\begin{itemize}
\item H2.1: Unimodal intent communication
\begin{itemize}
\item H2.1.1: The use of micromotion to indicate the user's order will improve user experience compared to the baseline condition in Part 1. 

\item H2.1.2: The use of visualizations to indicate the user's order  will improve user experience compared to the baseline condition in Part 1. 

\item H2.1.3: The use of speech to indicate the user's order  will improve user experience compared to the baseline condition in Part 1. 

\item H2.1.4: Among unimodal conditions, speech will yield the highest delivery accuracy and user experience.
\end{itemize}
\item H2.2: A multimodal combination (micromotion + visualization + speech) for order indication will yield better user experience and higher delivery accuracy than using any single modality alone. 

\end{itemize}

\begin{table}[h]
\centering
\setlength{\tabcolsep}{2pt}
\begin{tabular}{lcc}
\toprule
Metric & Baseline & Non-verbal Acoustic\\
\midrule
Work Fluency & 2.96 ± 1.23 & \textbf{3.38 ± 1.13}\\
Intelligence & 2.96 ± 1.37 & \textbf{3.33 ± 1.17}\\
Interaction Fluency & 2.83 ± 1.49 & \textbf{3.50 ± 1.06}\\
Reliability & 3.46 ± 1.14 & \textbf{3.67 ± 1.13}\\
Dependability & 3.33 ± 1.20 & \textbf{3.54 ± 1.28}\\
Goal Perception & 3.25 ± 1.36 & \textbf{3.67 ± 1.09}\\
Goal Communication & 3.04 ± 1.37 & \textbf{3.58 ± 1.14}\\
Collaboration & 3.12 ± 1.36 & \textbf{3.42 ± 1.21}\\
Satisfaction & 3.08 ± 1.14 & \textbf{3.62 ± 1.17}\\
Speed & 3.33 ± 1.17 & \textbf{3.79 ± 1.06}\\
\bottomrule
\end{tabular}
\caption{Mean and standard deviation of subjective ratings comparing the baseline condition and non-verbal acoustic cues for capturing user attention. No significant difference was found. Bold values indicate the higher mean.}
\label{tab:base_vs_sound}
\end{table}

\section{RESULTS}

We collected and analysed video recordings, surveys, and interview transcripts data from N=24 participants and conducted one-way ANOVA and post-hoc Tukey HSD in a pairwise comparision of all the modalities of interactions.

\subsection{\textbf{Part 1: Baseline vs non-verbal acoustic cues for user attention capturing}}
\subsubsection{\textbf{H1: The presence of a non-verbal acoustic cue while the robot approaches the user will improve user experience compared to having no cue.}}

To evaluate the hypothesis that the presence of a non-verbal acoustic cue while the robot approaches the user improves user experience compared to a baseline of no communicative cue, we analyzed the subjective questionnaire ratings. The results, presented in Table~\ref{tab:base_vs_sound}, show that mean scores for all metrics were higher in the `non-verbal acoustic cues' condition. However, an ANOVA showed no significant difference in any of these subjective ratings.


\subsection{\textbf{Part 2: Visualisation, micromotion, speech, and
multimodal intention communication}}

\begin{figure*}[h]
    \centering
    \subfloat[Percentage of participants thinking they picked correct cup vs actual correct cup]{\includegraphics[width=0.5\linewidth]{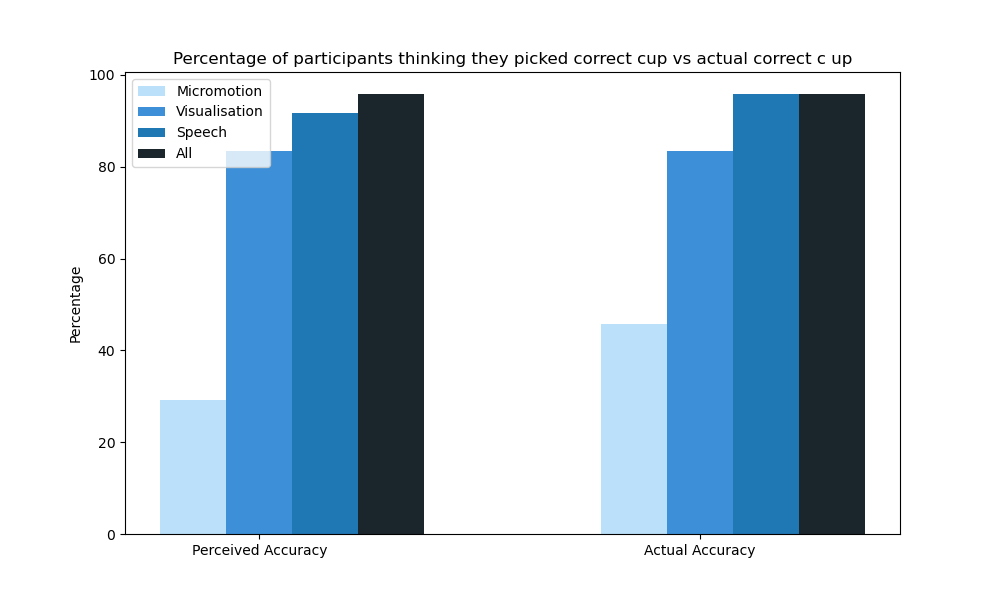}\label{fig:cue_comparison}}
    \hfill
    \subfloat[Total time taken to complete the delivery]
    {\includegraphics[width=0.5\linewidth]{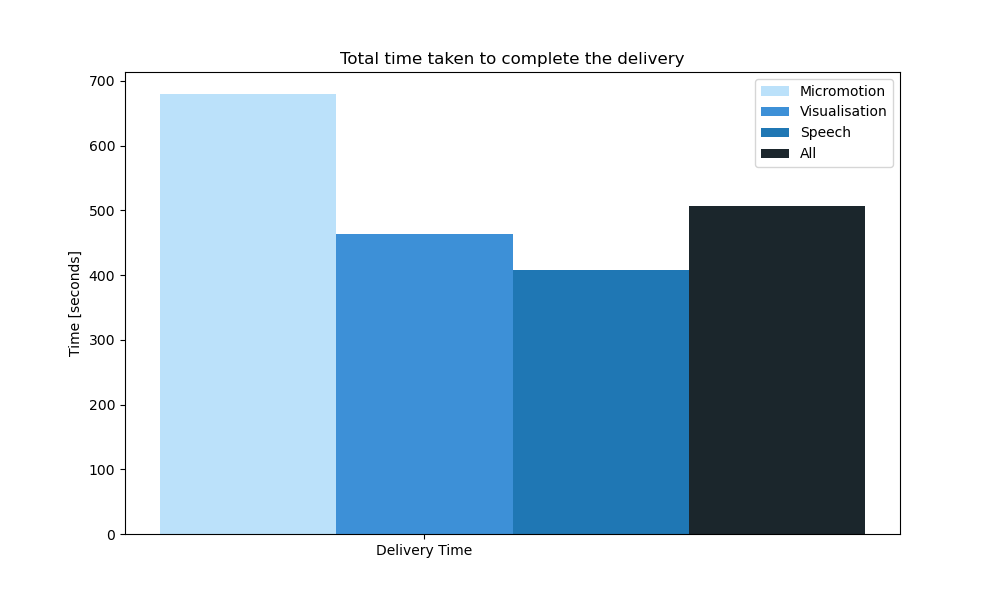}\label{fig:delivery_times}}
    \caption{Visualisation comparing Micromotion, Speech, Visualisation, and Multimodal conditions for 3 metrics: Perceived Accuracy [Pickup], Actual Accuracy [Pickup] and Delivery Time}
    \label{fig:delivery_accuracy}
\end{figure*}

\subsubsection{\textbf{H2.1.1: The use of micromotion to indicate the user's order will improve user experience compared to the baseline condition in Part 1.}}

\begin{table*}[h]
\centering
\begin{tabular}{lccccc}
\toprule
 Metric & Baseline & Micromotion & Visualisation & Speech & Multimodal \\
\midrule
Work Fluency & 2.96 ± 1.23 & 2.96 ± 1.33 & 3.50 ± 1.50 & \textbf{3.79 ± 1.25} & 3.62 ± 1.28 \\
Intelligence & 2.96 ± 1.37 & 3.00 ± 1.35 & 3.38 ± 1.41 & 3.62 ± 1.24 & \textbf{3.79 ± 1.06} \\
Interaction Fluency & 2.83 ± 1.49 & 2.83 ± 1.46 & 3.29 ± 1.49 & 3.62 ± 1.31 & \textbf{3.71 ± 1.30} \\
Reliability & 3.46 ± 1.14 & 3.12 ± 1.36 & 3.38 ± 1.47 & 3.62 ± 1.35 & \textbf{3.75 ± 1.15} \\
Dependability & 3.33 ± 1.20 & 3.00 ± 1.35 & 3.46 ± 1.47 & 3.58 ± 1.21 & \textbf{3.67 ± 1.13} \\
Goal Perception & 3.25 ± 1.36 & 2.96 ± 1.27 & 3.38 ± 1.47 & 3.46 ± 1.25 & \textbf{3.79 ± 1.18}\\
Goal Communication & 3.04 ± 1.37 & 2.92 ± 1.50  & 3.67 ± 1.49 & 4.12 ± 1.12 & \textbf{4.25 ± 0.94} \\
Collaboration & 3.12 ± 1.36 & 2.79 ± 1.32 & 3.54 ± 1.50 & 3.46 ± 1.35 & \textbf{3.92 ± 1.18} \\
Satisfaction & 3.08 ± 1.14 & 2.58 ± 1.32 & \textbf{3.62 ± 1.31} & 3.58 ± 1.18 & \textbf{3.62 ± 1.21} \\
Speed & 3.33 ± 1.37 & 3.38 ± 1.24 & 3.92 ± 1.25 & 3.79 ± 1.06 & \textbf{4.00 ± 0.88} \\
\bottomrule
\end{tabular}
\caption{Mean and Std of the subjective ratings comparing the unimodal and multimodal conditions for intention communication in Part 2. The baseline ratings from Part 1 are copied in this table for easy comparison. Multimodal communication received the highest subjective ratings, except for work fluency. Significant differences were found in baseline vs speech on work fluency (p = 0.025) and goal communication (p = 0.004), as well as baseline vs multimodal on intelligence (p=0.023), goal communication (p=0.001), collaboration (p=0.036). Further, multimodal is significantly better than micromotion on some of the metrics, while no significant difference was found in subjective ratings between multimodal and speech and visualisation conditions.}
\label{tab:uni vs multimodal}
\end{table*}

To assess the hypothesis that micromotions indicating the user's order enhance user experience compared to the baseline, we analyzed user experience metrics for the baseline and micromotion conditions, as presented in Table~\ref{tab:uni vs multimodal}. One-way ANOVA found no significant differences in participants' subjective perceptions between `base' and `micromotion', thus rejecting H2.1.1.

\subsubsection{\textbf{H2.1.2: The use of visualizations to indicate the user's order will improve user experience compared to the baseline condition in Part 1.}}

To test the hypothesis that visualizations indicating the user's order enhance user experience compared to the baseline, we analyzed user experience metrics for the `base' and `visualisation' conditions, as shown in Table~\ref{tab:uni vs multimodal}. Mean scores were consistently higher in the `visualisation' condition. However, one-way ANOVA found no significant differences between `base' and `visualisation', thus rejecting H2.1.2.



\subsubsection{\textbf{H2.1.3: The use of speech to indicate the user's order  will improve user experience compared to the baseline condition in Part 1.}}

To test the hypothesis that speech indicating the user's order enhances user experience compared to the baseline, we analyzed user experience metrics for the `base' and `speech' conditions, as shown in Table~\ref{tab:uni vs multimodal}. Mean scores were higher in the `speech' condition. One-way ANOVA indicated significant differences for `Temi and I Worked fluently together' (p = 0.025) and `I understood what Temi's goals were' (p = 0.004), but not for other metrics, partially supporting H2.1.3. 




\subsubsection{\textbf{H2.1.4: Among unimodal conditions, speech will yield the highest delivery accuracy and user experience.}}

To evaluate the hypothesis that speech performs best among unimodal conditions for order accuracy and user experience, we analyzed metrics comparing micromotion, visualisation, and speech, as shown in Tables~\ref{tab:uni vs multimodal}. Speech outperformed micromotion in user experience, with significant differences in fluency (p = 0.030), goal communication (p = 0.003), satisfaction (p = 0.008), and accuracy, with higher scores in cup selection (p = 0.009) and actual pickup accuracy (p = 0.000).

\subsubsection{\textbf{H2.2: A multimodal combination (micromotion + visualization + speech) for order indication will yield better user experience and higher delivery accuracy than using any single modality alone.}}

To test the hypothesis that a multimodal combination (micromotion, visualization, and speech) for order indication results in better user experience than any single modality alone, we analysed user experience metrics for the multimodal condition against speech, visualisation, and micromotion, as shown in Tables~\ref{tab:uni vs multimodal}.

The multimodal condition is significantly better than the baseline condition in intelligence (p=0.023), goal communication (p=0.001), and collaboration (p=0.036). 
The multimodal condition yielded higher mean scores than micromotion with significant differences in three metrics: Cup Selection (p=0.009), Perceived Accuracy (p=0.000) and Actual Accuracy (p=0.000), but showed no significant differences compared to speech and visualisation. 

Figure \ref{fig:cue_comparison} displays the percentage of ``Yes'' responses across Micromotion, Speech, Visualisation, and Multimodal conditions for two metrics: Perceived Accuracy and Actual Accuracy, supporting the hypothesis that a multimodal combination enhances order accuracy. The multimodal condition consistently achieves the highest rates—95.83\% across Perceived Accuracy and Actual Accuracy—outperforming Micromotion (e.g., 45.83\% for Actual Accuracy). The analysis of micromotion, speech, visualization and multimodal data revealed three statistically significant metrics (p < 0.05). Participants rated Temi's goal communication higher (4.12 ± 1.12 and 4.25 ± 0.94) compared to other metrics, with a significant F-value of 5.28 and p-value of 0.002, indicating strong differences across groups. Additionally collaboration showed significance (F = 2.92, p = 0.038), with mean scores ranging from 2.79 ± 1.32 to 3.92 ± 1.18. Satisfaction with Temi's service was also statistically significant (F = 4.03, p = 0.010), with scores varying from 2.58 ± 1.32 to 3.62 ± 1.21, reflecting notable differences in user satisfaction across the evaluated groups. The superior performance of the multimodal condition, likely due to participants having greater control over which cue they prefer, though non-significant p-values suggest variability in cue perception. Additionally, Figure \ref{fig:delivery_times} shows that the total time taken to complete the delivery is shortest for the speech condition.

\section{DISCUSSION}
\subsection{Major Findings and Implications}
\label{subsec:findings_implications}
This study highlights the critical need for efficient robot-customer interactions when users are preoccupied with cognitively demanding tasks, such as playing MonkeyType on their phones, simulating busy restaurant patrons. Effective interaction in such contexts requires two essential components: grabbing attention and communicating handoff intention. In Part 1, the baseline condition showed that 19/24 participants failed to notice the robot’s arrival [from post experiment interview], resulting in minimal interaction, as the robot simply waited with a single cup for 10 seconds before departing. In contrast, the non-verbal acoustic cues condition, using the EVA sound from \textit{WALL-E}, successfully captured participants’ attention, though the decision to pick up the cup depended on their willingness and availability, suggesting non-verbal acoustic cues’s efficacy in alerting distracted users. In Part 2, the visualisation condition was well-received, as participants could glance at the robot’s screen at their convenience, finding the instruction clear and unambiguous, aligning with prior work on visual cues~\cite{c2},~\cite{c6}. The speech condition required participants to remember verbal instructions until they chose to act, which was effective in our controlled setting but may pose challenges in real-world scenarios ~\cite{c9},~\cite{c10}. Notably, micromotion gestures (bowing and rotating to position the correct cup closer) were often unnoticed, and the final cup position was not consistently interpreted as indicating the correct choice, underscoring limitations in subtle motion cues for distracted users~\cite{c8},~\cite{c12}. Our findings, supported by significant improvements in order accuracy for the multimodal condition over micromotion (p $\leq$ 0.009) and visual indication in audio and motion indications (p $\leq$ 0.021) in Part 2, suggest that non-verbal acoustic cues and speech excel at grabbing attention, while visualisation is superior for communicating handoff intention. The multimodal combination leverages these strengths, enhancing overall interaction clarity and order accuracy, particularly for busy customers, by strengthening the vicarious sense of agency through combined cues~\cite{c11}.

\subsection{Limitations and Future Work}
\label{subsec:limitations_future_work}
Although our research offers significant insights, we did face certain limitations that necessitate further investigation. The use of MonkeyType to simulate a cognitively demanding task, though effective in engaging participants, creates a competitive environment that penalizes distractions, which may not align with real-world restaurant settings where minor interruptions (e.g., to retrieve food) are generally acceptable. Additionally, conducting experiments in a controlled lab environment may not fully capture the complexities of actual restaurant settings, including diverse customer behaviors and external factors. To address these issues, we plan to extend our research to real-world restaurant environments, testing robot intention communication methods with a larger and more diverse group of participants. Moreover, we tested a single robot platform in this study. Other robots with different appearances and communication modalities, such as a humanoid capable of hand gestures or an android with facial expressions, may result in different user preferences and communication effectiveness. Expanding the sample size is expected to enhance the likelihood of achieving statistically significant results. Participant feedback from open-ended survey responses and exit interviews indicates that visual cues could be improved, such as by adding LED displays on the robot’s tray to clearly indicate the correct order and reduce confusion~\cite{c2},~\cite{c10}. We also aim to improve interaction design by including user-driven inputs for personalized experiences, such as allowing repeated instruction requests to address challenges like directional dyslexia \footnote{Directional dyslexia refers to difficulty distinguishing left from right, a common perceptual challenge that can lead to errors in spatial interpretation, even among neurotypical individuals.} noted in the speech modality. Future work will explore conveying urgency through repeated cues or persistent prompts to engage busy customers. Additionally, we intend to investigate the impact of conflicting multimodal signals (e.g., one modality indicating one cup while another points to a different cup) on user perception and decision-making, potentially uncovering new dynamics in human-robot interaction (HRI) Additionally, social distance cues—such as the robot’s approach distance and orientation—were not varied or analyzed in this study. Although the robot maintained a consistent side approach across all conditions to minimize spatial confounds, prior work shows that approach direction and proximity significantly affect perceived threat and interaction willingness in seated, distracted users~\cite{c13}. In real restaurant settings, customers may respond differently to frontal versus side approaches, or to closer approaches during high cognitive load. Future work would systematically manipulate social distance to examine its interaction with multimodal intent cues and user preoccupation, potentially revealing trade-offs between attention capture and perceived invasiveness.

\section{CONCLUSIONS}

This study investigated the effectiveness of a multimodal approach (micromotion, visualization, and speech) for robot order indication with busy customers engaged in a cognitively demanding task, playing MonkeyType on their phones, simulating busy customers. Our findings from two studies demonstrate that effective robot-customer interaction requires both grabbing attention and communicating handoff intention, particularly as service robots become commonplace and customer engagement decreases~\cite{c13}. In Part 1 of our study, the baseline condition showed minimal interaction due to participants failing to notice the robot, while the non-verbal acoustic cues condition, using the EVA sound from \textit{WALL-E}, successfully alerted users, significantly improving interaction rates (p = 0.025 for fluency)~\cite{c4}. Part 2 of our study revealed that visualisation provided clear, user-friendly instructions, with high satisfaction (p = 0.008 vs. micromotion), while speech was effective but posed a challenge for a user with directional dyslexia ~\cite{c9},~\cite{c10}. Micromotion gestures were often unnoticed, limiting their effectiveness (p $\leq$ 0.009 for multimodal vs. micromotion accuracy). The multimodal combination outperformed micromotion in most accuracy metrics (p $\leq$ 0.009) and visualisation in audio and motion indications (p $\leq$ 0.021), partially supporting our hypothesis that it achieves higher order accuracy than single modalities alone. These results suggest that non-verbal acoustic cue and speech excel at capturing attention, while visualisation is optimal for communicating intent, with the multimodal approach enhancing overall clarity by strengthening the vicarious sense of agency~\cite{c11}. As robots transition from novel to routine in service settings, our findings highlight the potential of multimodal cues to ensure efficient interactions with increasingly busy customers, paving the way for real-world applications in dynamic restaurant environments.





\end{document}